%% file: ms.tex
\documentclass[submission,copyright,creativecommons,a4paper,superscriptaddress,shorttitle=Functorial Language Games for QA]{eptcs}
\pdfoutput=1
\providecommand{\event}{Applied Category Theory 2020}

\usepackage[utf8]{inputenc}

\input{header}

\input{macros}

\input{gamediagrams}

\DeclareRobustCommand{\coprod}{\mathop{\text{\fakecoprod}}}
\newcommand{\fakecoprod}{
  \sbox0{$\prod$}
  \smash{\raisebox{\dimexpr.9625\depth-\dp0}{\scalebox{1}[-1]{$\prod$}}}
  \vphantom{$\prod$}
}

\title{Functorial\ Language\ Games\ for\ Question\ Answering}

\author{Giovanni de Felice\texorpdfstring{$^\dagger$}{}, Elena Di Lavore\texorpdfstring{$^\star$}{}, Mario Román\texorpdfstring{$^\star$}{}, Alexis Toumi\texorpdfstring{$^\dagger$}{}
  \institute{$\dagger$ Department of Computer Science, University of Oxford.\\ $\star$ Department of Software Science, Tallinn University of Technology.}}

\def\titlerunning{Functorial Language Games for QA}
\def\authorrunning{de Felice, Di Lavore, Román, Toumi}

\begin{document}

\maketitle

\makeatletter{\renewcommand*{\@makefnmark}{}
\footnotetext{Elena Di Lavore and Mario Román were supported by the European Union through the ESF funded Estonian IT Academy research measure (project 2014-2020.4.05.19-0001).}\makeatother}

\begin{abstract}
  We present some categorical investigations into Wittgenstein's
  language-games, with applications to game-theoretic pragmatics and
  question-answering in natural language processing.
\end{abstract}

\section*{Introduction}\input{0-intro}
\section{Q\&A as an open game}\label{sec:open-games}\input{1-open-games}
\section{Pregroup semantics}\label{sec:pregroup-semantics}\input{2-pregroup}
\section{From pregroups to open games with snake removal}\label{sec:pregroup-games}\input{3-grammar-to-games}
\section{Functorial language games}\label{sec:functorial-language}\input{4-functorial-language}
\section{Nash equilibria in a Q\&A game}\label{sec:nash-equilibria}\input{5-nash-equilibria}

\section{Conclusion}
We studied the links between two recently developed applications of category theory: open games and distributional compositional language models.
We constructed language games as functors from a pregroup grammar to the
free completion of open games as a rigid category and used this construction
to give a game-theoretic pragmatics for orders and questions.
Finally, we analysed the Nash equilibria of an adversarial Q\&A game.
Going towards implementation, the next step is to define a similar Q\&A game in
the category of learners of Fong et al. \cite{fong2019}, see \cite{hedges2019}
for the relationship between learners and open games.
This would amount to instantiating the strategy sets with the parameters of a
learning algorithm, so that gradient descent converges to the desired Nash
equilibrium. It would allow to formalise a generative adversarial learning
algorithm \cite{goodfellow2014a} for question answering.

\footnotesize
\bibliographystyle{eptcs}
\bibliography{main}

\end{document}

%% file: header.tex
\usepackage[utf8]{inputenc}
\usepackage[english]{babel}
\usepackage[T1]{fontenc}
\usepackage{hyperref}

\usepackage{mathtools}
\usepackage{amsthm}
\usepackage{amsmath}
\usepackage{amssymb}
\usepackage{mathrsfs}
\usepackage{stmaryrd}
\usepackage{braket}
\usepackage{physics}
\usepackage{tikz}
\usetikzlibrary{matrix,arrows,decorations.pathmorphing}
\usetikzlibrary{decorations.markings}
\usetikzlibrary{shapes.geometric}

\newcommand{\s}{\enspace}
\newcommand{\sub}{\subseteq}
\newcommand{\size}[1]{\left\vert{#1}\right\vert}
\renewcommand{\tt}[1]{\mathtt{#1}}
\renewcommand{\bf}[1]{\ensuremath{\mathbf{#1}}}

\renewcommand{\cal}[1]{\mathcal{#1}}
\newcommand{\bb}[1]{\mathbb{#1}}
\renewcommand{\phi}{\varphi}

\newcommand{\injects}{\xhookrightarrow{}}

\newtheorem{definition}{Definition}[section]
\newtheorem{proposition}[definition]{Proposition}

\newtheorem{lemma}[definition]{Lemma}

\newtheorem{example}[definition]{Example}

\usepackage{tabularx,environ}
\makeatletter
\newcommand{\problemtitle}[1]{\gdef\@problemtitle{#1}}
\newcommand{\probleminput}[1]{\gdef\@probleminput{#1}}
\newcommand{\problemoutput}[1]{\gdef\@problemoutput{#1}}
\NewEnviron{problem}{
  \problemtitle{}\probleminput{}\problemoutput{}
  \BODY

   \@problemtitle
  \par\addvspace{.5\baselineskip}
  \noindent
  \begin{tabularx}{\textwidth}{@{\hspace{\parindent}} l X c}
    \textbf{Input:} & \@probleminput \\
    \textbf{Output:} & \@problemoutput
  \end{tabularx}
  \par\addvspace{.5\baselineskip}
}

%% file: macros.tex
\newcommand\G{\mathcal{G}}
\renewcommand\Set{\mathit{Set}}

\usepackage{tikz}
\usepackage{varwidth}
\usetikzlibrary{decorations.markings}
\tikzset{->-/.style={decoration={markings, mark=at position #1 with {\arrow{>}}}, postaction={decorate}},->-/.default=0.5}
\tikzset{game/.style={rectangle, minimum height = #1, minimum width = .8 cm, draw},game/.default=1.2cm}
\tikzset{dot/.style={circle, scale=.5, fill=#1, draw},dot/.default=black}
\tikzset{player/.style={isosceles triangle, isosceles triangle apex angle=90, inner sep=0pt, minimum width=2cm, shape border rotate=#1, draw},player/.default=180}
\newcommand{\diset}[2]{\binom{#1}{#2}}
\newcommand{\game}[6]{#1 \colon  \diset{#2}{#3} \overset{#6} \nrightarrow \diset{#4}{#5}}

\newcommand{\Pf}{\pi}
\newcommand{\Cf}{\kappa}
\newcommand{\Ef}{E}

\newcommand{\amax}{\operatorname*{argmax}}

\newcommand\teacher{\mathcal{T}}
\newcommand\student{\mathcal{S}}
\newcommand\marker{\mathcal{M}}
\newcommand\dist{d}

%% file: gamediagrams.tex
\newcommand\teacherstudentdiagram{
\begin{tikzpicture}
\node(l1)at(0,.8){};
\node(l2)at(0,.5){};
\node(l3)at(0,.3){};
\node(l4)at(0,-.3){};
\node(l5)at(0,-.8){};
\node(l6)at(0,-.8){};
\node[game](T)at(1,0){$\mathcal{T}$};
\node[game](S)at(3,0){$\mathcal{S}$};
\node[player=0](M)at(5,0){$\mathcal{M}$};
\node[isosceles triangle, isosceles triangle apex angle=90, inner sep=0pt, minimum width=1.2cm, shape border rotate=180, draw](c)at(-.5,|-T){$f$};
\draw[->-](c.east)to node[above]{$C$}(T.west);
\draw[->-](T.east)[out=0,in=180]to (S.west)node [above left]{$Q$};
\draw[->-](T.east|-l3)[out=0,in=180]to(3,|-l1) node [above]{$A$}to(M.west|-l1);
\draw[->-](S.east|-l3)[out=0,in=180]to node [above]{$A$}(M.west|-l3);
\draw[->-](M.west|-l4)[out=180,in=0]to node [below]{$U$}(S.east|-l4);
\draw[->-](M.west|-l5)[out=180,in=0]to(3,|-l6) node [below]{$U$}[out=180,in=0]to(T.east|-l4);
\end{tikzpicture}
}

%% file: 0-intro.tex

In his 1953 \emph{Philosophical Investigations} \cite{wittgenstein1953}, Wittgenstein
introduces the concept of \emph{language-game} (\emph{Sprachspiel}) as a basis for his theory of meaning.
He never gives a general definition, and instead proceeds by enumeration of
examples: ``asking, thanking, cursing, greeting, praying''.
Thus, depending on the language-game in which it is played, the same utterance
``Water!'' can be the answer to a question, a request to a waiter or the chorus
of a song. This has often been summarised by the slogan ``meaning is use'',
which became the object of a subfield of linguistics: \emph{pragmatics}.
Since Lewis' work on language conventions \cite{lewis1969}, formal game
theory has been used to model speaker's and hearer's actions
in the context of a discourse~\cite{benz2018}. In parallel, game theory
has been proven significant in designing machine learning tasks \cite{goodfellow2014a} and is beginning to be applied to natural language
processing \cite{subramanian2017, tripodi2019}.

Category theory has been used to formalise both language and games.
On the one hand, the \emph{distributional compositional} (DisCo) models of Coecke et al. \cite{ClarkEtAl08, ClarkEtAl10} define language meaning as a functor from Lambek's pregroup grammars \cite{Lambek99} to compact-closed categories such as the category of vector spaces and linear maps.
On the other hand, Ghani et al. \cite{GhaniHedges18} introduced a monoidal category of \emph{open games} as a compositional framework for game theory.
In \cite{HedgesLewis18}, Hedges and Lewis proposed to construct language games as functors from grammar to open games.
However, their construction required to depart from the pregroup formalism and relied on an open conjecture.

This paper presents some categorical investigations into language games, with applications to game-theoretic pragmatics.
In the first two sections, we give an abstract definition of question-answering as an open game, then define pregroup grammars and DisCo models.
The proposal of Hedges and Lewis is reformulated within the pregroup formalism,
using the free completion of open games as a rigid monoidal category.
Two examples of functorial language games are presented: orders and questions,
modelling the pragmatics of imperative and interrogative syntax.
The abstract question-answering game of the first section is instantiated
with respect to a pregroup grammar: the teacher's moves are grammatical questions,
the student's strategies are DisCo models.
We characterise the Nash equilibria and identify sufficient conditions
for the student to succeed.
We conclude with a discussion of the link between open games and learning algorithms
for natural language processing.

%% file: 1-open-games.tex

Open games~\cite{GhaniHedges18} are building blocks that can be
composed together to construct games in the sense of economic game
theory.  They provide a compositional description of game theory and a
graphical syntax that facilitates reasoning. They express
the equilibria of composite games in terms of their components, which
might be more tractable.

\begin{definition}[{{\cite[Definition 3]{GhaniHedges18}}}]
  Let $X,S,Y,R \in \Set$. An \emph{open game} $\game{\G}{X}{S}{Y}{R}{\Sigma}$ that
  takes \emph{observations} on the set $X$, produces \emph{moves} on the set
  $Y$, receives \emph{utilities} on the set $R$, and returns \emph{coutilities}
  on the set $S$, is a quadruple $(\Sigma_\G, \Pf_\G, \Cf_\G, \Ef_\G)$ where:
  \begin{itemize}
    \item $\Sigma_\G$ is the set of \emph{strategy profiles};
    \item $\Pf_\G \colon \Sigma_\G \times X \to Y$ is the \emph{play function},
    representing how the player following a strategy profile produces a move
    given an observation;
    \item $\Cf_\G \colon \Sigma_\G \times X \times R \to S$ is the
    \emph{coplay function}, representing how payoffs are propagated from the
    player to its environment;
    \item $\Ef_\G \colon X \times (Y \to R) \to \mathcal{P}(\Sigma_\G)$ is the
    \emph{equilibrium function}, representing the subset of strategies that are
    \emph{best responses} for the player in a given context, which consists of a
    past observation of type $X$ and a continuation of type $Y \to R$.
  \end{itemize}
\end{definition}

Open games form a teleological symmetric monoidal category denoted $\bf{Game}$, which admits a graphical calculus
developed in~\cite{Hedges17}. Each morphism is represented as a box with
covariant wires for observations and moves, and contravariant wires for
utilities and coutilities (see Diagram \ref{diagram_teacher_student}).
\emph{Closed} games $\G \colon I \to I$ are determined by a set of strategy profiles $\Sigma_\G$ together with an equilibrium function
$\Ef_\G \colon 1 \to \mathcal{P}(\Sigma_\G)$ describing a
subset of \textit{equilibria}. When considering games where players maximise their own utility, these equilibria are precisely \textit{Nash
equilibria}~\cite[Theorem 2]{GhaniHedges18}. Every simultaneous move game,
in the sense of classical game theory, defines a closed game~\cite[Section VII]{GhaniHedges18}.
Finally, every pair of play/coplay functions can be lifted into a game by picking the
trivial strategy set $\bf{S}_{\mathcal{G}} = 1$ and a constantly true equilibrium function.
Open games obtained in this way are called \textit{strategically trivial}.

\begin{example}\label{ex:one-shot-game}
  Consider an agent $q$ that poses questions, modelled as a state
  $\game{q}{1}{1}{Q}{U}{}$ with a set of strategies given by $Q$. We pick some
  questions-answer pairs $T \subseteq Q \times U$ that the agent finds
  \emph{satisfactory}, and these preferences are modelled by defining its
  equilibrium function to be
  \[
    \Ef_{\mathcal{G}}(f) \coloneqq \left\{ q \mid (q , f(q)) \in T\right\}.
  \]
  It confronts some oracle that answers the questions, modelled as a
  strategically trivial effect $\game{a}{Q}{U}{1}{1}{}$.
  Composing both gives a simple game where the agent tries to ask a question
  and receive a satisfactory answer.
  The Nash equilibria of the composite game
  are precisely the questions that the oracle answers satisfactorily.
\end{example}

Generalizing from this example, we can consider an abstract notion of
\emph{utility-maximising player}.

\begin{definition}\label{def:utility-maximising}
  A \emph{utility-maximising} player with observations
  in $X$, moves in $Y$, utilities in a partially ordered set $R$, and some
  subset of possible strategies $\Sigma \subseteq Y^{X}$, is an open game
  $\game{\G}{X}{1}{Y}{R}{\Sigma}$ with $\Pf_\G (\sigma, x) \coloneqq \sigma (x)$
  the evaluation function, trivial coplay function; and
  $\Ef_{\G}(x,\kappa) \coloneqq \amax_{\sigma \in \Sigma} \kappa (\sigma (x))$
  an equilibrium function describing the subset of the strategies that maximises
  the utility of the agent.
\end{definition}

Let us fix three sets $C$, $Q$, $A$ for corpora (i.e. lists of facts),
questions and answers respectively. Let $U$ be a set of utilities, which
can be taken to be $\bb{R}$ or $\bb{B}$.
We now define a closed game modelling an interaction between three
agents: a teacher, a student and a marker.  For the moment
we define it in terms of abstract sets and unspecified strategies, we will
instantiate these in Section~\ref{sec:nash-equilibria}.

\begin{example}[Teacher, student, marker]\label{def:qa-game}
A \textit{teacher} $\game{\teacher}{C}{1}{Q \times A}{U}{}$ is a utility-ma\-xi\-mi\-sing player
where each strategy represents a
function turning facts from the corpus into pairs of questions and answers.
A \textit{student} $\game{\student}{Q}{1}{A}{U}{}$ is a utility-maximising player
where each startegy represents a way of
turning questions into answers.
A \textit{marker} is a strategically trivial open game
$\game{\marker}{A \times A}{U \times U}{1}{1}{}$
with trivial play function and a coplay function defined as
$\Cf_\marker (a_\teacher, a_\student) = (-\dist(a_{\student}, a_{\teacher}),
\dist(a_{\student}, a_{\teacher}))$
where $\dist: A \times A \to U$ is a given metric on $A$.
Finally, we model a \textit{corpus} as a strategically trivial
open game $\game{f}{1}{1}{C}{1}{}$ with play function given by
$\Pf_f (*) = f \in C$.
All these open games are composed to obtain a \textit{question answering game}
in the following way.
\begin{equation}
  \label{diagram_teacher_student}
  \teacherstudentdiagram
\end{equation}
Intuitively, the teacher produces a question from the corpus and gives it
to the student who uses his strategy to answer. The marker will receive
the correct answer from the teacher together with the answer that the student produced,
and output two utilities. The utility of the teacher will be the distance
between the student's answer and the correct answer; the utility of the
student will be the exact opposite of this quantity. In this sense,
question answering is a zero-sum game.
\end{example}

%% file: 2-pregroup.tex
Pregroup grammars are algebraic models of natural language syntax, first
introduced by Lambek \cite{Lambek99}. They are weakly equivalent to context-free
grammars \cite{buszkowski2007}, can be parsed efficiently \cite{preller2007},
and can be given functorial semantics \cite{ClarkEtAl10, sadrzadeh2013}.
In this section we give background on pregroups and their models in rigid
monoidal categories.

\begin{definition}
A \emph{preordered monoid} is a preorder $P$ equipped with a monotone monoid,
i.e. $a \leq c \s \land \s b \leq d \implies a b \leq c d$
for all $a, b, c, d \in P$.
A \emph{pregroup} is a preordered monoid $P$ where every type $t \in P$ has left
and right adjoints $t^l, t^r$ such that $t^l t \leq 1 \leq t t^l$ and
$t t^r \leq 1 \leq t^r t$.
\end{definition}

\begin{definition}
  A \emph{pregroup grammar} is a tuple $G = (V, B, D, s)$ where $V$ is a set of words called the \emph{vocabulary}, $B$ is a finite set of
  \emph{basic types} with $s \in B$ the \emph{sentence type}, and
  $D \sub V \times P_B$ is a finite set of \emph{dictionary entries}
  for $P_B$ the free pregroup generated by $B$.

  A pregroup grammar $G$ generates a language $\cal{L}(G, s) \sub V^\ast$
  as follows. A list of words $u \in V^\ast$ is grammatical, i.e.
  $u \in \cal{L}(G, s)$, whenever for each word $u_i \in V$, $i \leq \size{u}$
  there is a type $t_i \in P_B$ such that $(u_i, t_i) \in D$ and
  $t_1 \dots t_n \leq s$ in the free pregroup.
\end{definition}

In \cite{PrellerLambek07}, Lambek and Preller recast the pregroup formalism
in terms of \emph{free compact 2-categories}. We focus on the case of
compact 2-categories with one-object, i.e. \emph{rigid categories}.
Let $\bf{Cat}$, $\bf{MonCat}$ and $\bf{RigidCat}$ denote the categories of
categories and functors, monoidal categories and monoidal functors and rigid categories and rigid functors, respectively.
Given a set of generating objects $O$, a \emph{simple signature} over $O$
is a graph $\Gamma \rightrightarrows O$, it generates the free category
$\bf{C}(\Gamma) \in \bf{Cat}$, \cite[2.3]{Selinger09}.
A \emph{monoidal signature} over $O$ is a graph
$\Gamma \rightrightarrows O^\ast$ and it generates the free monoidal category
$\bf{MC}(\Gamma) \in \bf{MonCat}$, \cite[3.3]{Selinger09}.
Finally a \emph{rigid signature} over $O$
is a graph $\Gamma \rightrightarrows P_O$ where $P_O$ is the free pregroup on
$O$, $\Gamma$ generates the free rigid category
$\bf{RC}(\Gamma) \in \bf{RigidCat}$, \cite[4.8]{Selinger09}.

For a pregroup grammar $G = (V, B, D, s)$, let
$\bf{G} := \bf{RC}(D \rightrightarrows B + V)$ be the free rigid category
generated by the dictionary entries, where the two maps are given by
projections, i.e. $dom(w, t) = w$ and $cod(w, t) = t$ for $(w, t) \in D$.
The grammatical structure of a sentence $u \in \cal{L}(G, s)$ may now
be given explicitly by a diagram $g : u \to s$ in $\bf{G}$.
In general, for any basic type $b \in B$ we will write:
$$ \cal{L}(G, b) = \coprod_{u \in V^\ast} \bf{G}(u, b)$$

\begin{example}
    Take $G = (V, B, D, s)$ with $V = \set{\rm{sense}, \rm{makes}, \rm{this},
    \rm{sentence}}$, $B = \set{s, n, d}$ and $D = \set{(\text{sense}, n), \s (\text{makes}, n^r s n^l), \s (\text{this}, d), \s (\text{sentence}, d^l n)}$.
    The following string diagram is a proof that $u =$
    ``This sentence makes sense'' $\in \cal{L}(G, s)$ is a grammatical
    sentence. We do not draw the wires for words and depict the dictionary entries as triangles.
    \begin{center}
        \input{figures/makes_sense.tex}
    \end{center}
\end{example}

Going from inequalities in a preordered monoid to arrows in a monoidal category
allows both to reason about syntactic ambiguity (e.g. ``men and (women who run)''
vs ``(men and women) who run'') as well as to define pregroup semantics as a
monoidal functor. This second observation lead to the development of the DisCo
(distributional compositional) framework \cite{ClarkEtAl08,ClarkEtAl10}.

\begin{definition}
    A DisCo model for a pregroup grammar $G = (V, B, D, s)$ is a rigid monoidal
    functor $F : \bf{G} \to \bf{S}$ for $\bf{S}$ a rigid monoidal category,
    such that words are sent to the unit $F(w) = 1$ for all $w \in V$.
    The semantics of a list of words $u \in V^\ast$ with grammatical structure
    $g : u \to t$ is given by the state $F(g) : 1 \to F(t)$.
\end{definition}

\begin{example}\label{example:queries}
  Relational models $F: \bf{G} \to \bf{Rel}$ correspond precisely to relational
  databases. The basic types $B$ correspond to a set of \emph{attributes}, with
  $F(b)$ the set of \emph{data values} for each $b \in B$.
  The dictionary $D \sub V \times P_B$ corresponds to a \emph{database schema},
  and the functor $F$ to an \emph{instance} of that schema, mapping
  each entry $(w, t) \in D$ to a relation $F(w) \sub F(t) =
  F(b_0) \times F(b_1) \dots \times F(b_k)$ where $\{ b_i\}$ is the list of
  basic types in $t$ (note that $F(b^r) = F(b) = F(b^l)$ as $\bf{Rel}$
  is compact closed). Finally, morphisms in $\bf{G}$ (and sentences $u \to s$ in
  particular) correspond to \emph{conjunctive queries} which can be evaluated by
  applying the functor $F$, see \cite{DeFeliceEtAl19a} where this correspondence
  is spelled out in detail.
\end{example}

\begin{example}
  Distributional models $F:\bf{G} \to \bf{Vect}_\bb{R}$ can be constructed by
  counting co-occurences of words in a corpus \cite{Grefenstette11}. The
  image of the noun type $n \in B$ is a vector space where the inner product
  computes noun-phrase similarity \cite{sadrzadeh2013}.
  When applied to question answering tasks, distributional models can be used
  to compute the distance between a question and its answer \cite{coecke2018b}.
\end{example}

%% file: figures/makes_sense.tex
\begin{tikzpicture}[scale=0.85]
\node () at (1.25, -0.25) {$d$};
\draw (0.0, 0) -- (2.0, 0) -- (1.0, 1) -- (0.0, 0);
\node () at (1.0, 0.25) {This};
\node () at (3.4166666666666665, -0.25) {$d^r$};
\node () at (4.083333333333333, -0.25) {$n$};
\draw (2.5, 0) -- (4.5, 0) -- (3.5, 1) -- (2.5, 0);
\node () at (3.5, 0.25) {sentence};
\node () at (5.75, -0.25) {$n^r$};
\node () at (6.25, -0.35) {$s$};
\node () at (6.85, -0.25) {$n^l$};
\draw (5.0, 0) -- (7.0, 0) -- (6.0, 1) -- (5.0, 0);
\node () at (6.0, 0.25) {makes};
\node () at (8.75, -0.25) {$n$};
\draw (7.5, 0) -- (9.5, 0) -- (8.5, 1) -- (7.5, 0);
\node () at (8.5, 0.25) {sense};
\draw [out=-90, in=180] (1.0, 0) to (2.083333333333333, -1);
\draw [out=-90, in=0] (3.1666666666666665, 0) to (2.083333333333333, -1);
\draw [out=-90, in=180] (3.833333333333333, 0) to (4.666666666666666, -1);
\draw [out=-90, in=0] (5.5, 0) to (4.666666666666666, -1);
\draw [out=-90, in=180] (6.5, 0) to (7.5, -1);
\draw [out=-90, in=0] (8.5, 0) to (7.5, -1);
\draw [out=-90, in=90] (6.0, 0) to (6.0, -1.25);
\end{tikzpicture}

%% file: 3-grammar-to-games.tex
We now aim to give semantics to pregroup grammars in the category
of open games. As $\bf{Game}$ is not a rigid monoidal category, building
such an interpretation is more involved than the cases above.
A first attempt would be to use the teleological
structure of $\bf{Game}$ to interpret the cups in pregroup diagrams.
However, the models obtained in this way would be very limited: the image of
words are all states, i.e. they have trivial observations and co-utilities.
This motivated Hedges and Lewis \cite{HedgesLewis18} to define the notion of a
\emph{process grammar} as a free rigid category where the words are modeled
as generators with arbitrary domains.
We propose an alternative construction which does not require any change to
the pregroup formalism, using the following lemma.

\begin{lemma}[\cite{Delpeuch14a}]
  The forgetful functor $\bf{RigidCat} \to \bf{MonCat}$ has a left adjoint
  $\cal{A}: \bf{MonCat} \to \bf{RigidCat}$.
  Furthermore, the embedding functor $\bf{C} \injects \cal{A}(\bf{C})$ is strong
  monoidal and fully-faithful.
\end{lemma}

The free completion $\cal{A}: \bf{MonCat} \to \bf{RigidCat}$ is called
\emph{autonomisation} in \cite{Delpeuch14a}, given a monoidal category $\bf{C}$
it constructs a rigid monoidal category $\cal{A}(\bf{C})$ by freely adding
adjoints to the objects of $\bf{C}$ with formal cups and caps witnessing the
adjunctions.

In our context, this means that we can give semantics to pregroup grammars in
open games by constructing a DisCo model $F : \bf{G} \to \cal{A}(\bf{Game})$
with $F(s) \in \bf{Game}$. Then, by fullness of the embedding
$\bf{Game} \injects \cal{A}(\bf{Game})$, we get that the image of any
grammatical sentence $g: u \to s$ in $\bf{G}$ is a morphism:
$$F(g) \in \cal{A}(\bf{Game})(1, F(s)) \simeq \bf{Game}(1, F(s))\, .$$
In other words, once a functor of this type is constructed, all the cups and
caps from pregroup diagrams will cancel each other via the snake equation,
leaving us with a morphism in $\bf{Game}$.
We now give the main result of this section, which will allow us to build
language games functorially from any \emph{functional} pregroup grammar.

\begin{definition}
  A pregroup type $t \in P_B$ is \emph{functional} if it belongs to the
  context-free grammar $ b \: \vert \: t^r t \: \vert \: t t^l$ where $b \in B$.
  A pregroup grammar $G = (B, V, D, s)$ is called \emph{functional} if
  each type in the dictionary $t \in D(V)$ is functional.
\end{definition}

The restriction to functional types is very common in the literature on
categorial grammar. They correspond to the syntactic types of combinatory categorial grammars \cite{steedman2000} and of the \emph{product-free} Lambek
calculus \cite{fowler2008, buszkowski2016}, as well as the original types used
for Montague semantics \cite{Montague70a}. Even if Lambek doesn't state this
restriction explicitly, all the pregroup types he uses in his book
\cite{Lambek08} are of this form.

\begin{proposition}\label{prop:factorisation}
  For any functional pregroup grammar $G$, there exists a monoidal signature
  $\Gamma_G$ and a functor $F_G: \bf{G} \to \bf{RC}(\Gamma_G)$, such that
  any DisCo model $F: \bf{G} \to \cal{A}(\bf{Game})$ factors uniquely as
  $F = \cal{A}(J) \circ F_G$ for some $J: \bf{MC}(\Gamma_G) \to \bf{Game}$.
\end{proposition}

\begin{example}\label{ex:autonomisation}
  Fix a pregroup grammar $G = (B, V, D, s)$ with
  $B = \{ d, n, s\}$ for determinant, noun, sentence types. We use the dictionary:
  $$
  D = \{(\text{the},\, d), \; (\text{person},\, d^r n),\;
  (\text{who},\, n^r n s^l n),\;
  (\text{explains},\, n^r s), \; (\text{knows},\, n^r s n^l),\;
  (\text{rules},\, d^r n)\}.
  $$
  The following is a grammatical sentence.
  \begin{equation}\label{ex:sentence}
    \input{figures/sentence.tex}
  \end{equation}
  The monoidal signature $\Gamma_G$ associated with $G$ is given by:
  $\Gamma_G = \{ \text{the} : 1 \to d\, ; \,\text{person, rules}:d \to n\, ;
   \, \text{explains}: n \to s\, ;
   \text{knows}: n \otimes n \to s\, ;\, \text{who}_1: n \to a \otimes n\, ;
   \, \text{who}_2: a \otimes s \to n\}$.
  The functor $F_G$ factors each dictionary entry using cups and the boxes in
  $\Gamma_G$. For instance every entry of the form $(\text{w},\, a^r b) \in D$
  (including adjectives, common nouns and intransitive verbs) and every
  transitive verb $(\text{v},\, n^r s n^l) \in D$ are factored as follows:
  \begin{equation}
    \begin{gathered}
    \input{figures/itverbs.tex}
    \end{gathered}
    \quad , \quad\quad
    \begin{gathered}
    \input{figures/tverbs.tex}
    \end{gathered}
    .
  \end{equation}
  The factorization of ``who'' requires two boxes:
  \begin{equation}
    \begin{gathered}
     \input{figures/relpron.tex}
    \end{gathered}.
  \end{equation}
  Note that this factorisation carries the same data as a \emph{comb} in the sense
  of \cite{kissinger2017} or equivalently of a \emph{lens}, see \cite{roman2020}
  for an account of the relationship between these notions.
  Applying the functor $F_G$ to the sentence (\ref{ex:sentence})
  and removing the snakes yields the following diagram in
  $\bf{MC}(\Gamma_G) \injects \bf{RC}(\Gamma_G)$.
  \begin{equation*}
    \input{figures/F_G_sentence.tex}
  \end{equation*}
\end{example}

We can now build games for each sentence by finding a functor
$J: \bf{MC}(\Gamma_G) \to \bf{Game}$.
In the next section we will use this to obtain a game-theoretic semantics for
\emph{imperative} and \emph{interrogative} sentences.

%% file: figures/sentence.tex
\begin{tikzpicture}[scale=0.8]
\draw (0.0, 0) -- (2.0, 0) -- (1.0, 1) -- (0.0, 0);
\node () at (1.0, 0.25) {The};
\draw (2.5, 0) -- (4.5, 0) -- (3.5, 1) -- (2.5, 0);
\node () at (3.5, 0.25) {person};
\draw (5.0, 0) -- (7.0, 0) -- (6.0, 1) -- (5.0, 0);
\node () at (6.0, 0.25) {who};
\draw (7.5, 0) -- (9.5, 0) -- (8.5, 1) -- (7.5, 0);
\node () at (8.5, 0.25) {explains};
\draw (10.0, 0) -- (12.0, 0) -- (11.0, 1) -- (10.0, 0);
\node () at (11.0, 0.25) {knows};
\draw (12.5, 0) -- (14.5, 0) -- (13.5, 1) -- (12.5, 0);
\node () at (13.5, 0.25) {the};
\draw (15.0, 0) -- (17.0, 0) -- (16.0, 1) -- (15.0, 0);
\node () at (16.0, 0.25) {rules.};
\begin{scope}[yscale=0.7]
  \draw [out=-90, in=180] (1.0, 0) to (2.083333333333333, -1);
  \draw [out=-90, in=0] (3.1666666666666665, 0) to (2.083333333333333, -1);
  \draw [out=-90, in=180] (3.833333333333333, 0) to (4.616666666666667, -1);
  \draw [out=-90, in=0] (5.4, 0) to (4.616666666666667, -1);
  \draw [out=-90, in=180] (6.6, 0) to (7.383333333333333, -1);
  \draw [out=-90, in=0] (8.166666666666666, 0) to (7.383333333333333, -1);
  \draw [out=-90, in=180] (13.5, 0) to (14.583333333333332, -1);
  \draw [out=-90, in=0] (15.666666666666666, 0) to (14.583333333333332, -1);
  \draw [out=-90, in=180] (6.2, 0) to (7.5166666666666675, -2);
  \draw [out=-90, in=0] (8.833333333333334, 0) to (7.5166666666666675, -2);
  \draw [out=-90, in=180] (11.5, 0) to (13.916666666666666, -2);
  \draw [out=-90, in=0] (16.333333333333332, 0) to (13.916666666666666, -2);
  \draw [out=-90, in=180] (5.8, 0) to (8.15, -3);
  \draw [out=-90, in=0] (10.5, 0) to (8.15, -3);
  \draw [out=-90, in=90] (11.0, 0) to (11.0, -3.5);
\end{scope}
\end{tikzpicture}

%% file: figures/itverbs.tex
\begin{tikzpicture}
\node () at (0.25, 0.5) {$a^r$};
\node () at (1.25, 0.5) {$b$};
\draw [out=-90, in=90] (0.0, 0.75) to (0.0, 0.0);
\draw [out=-90, in=90] (1.0, 0.75) to (1.0, 0.0);
\draw (-0.25, 0.75) -- (1.25, 0.75) -- (0.5, 1.5) -- (-0.25, 0.75);
\node () at (0.5, 1.0) {w};
\node () at (2.0, 1.0) {$\mapsto$};
\node () at (3.25, 1.0) {$a^r$};
\node () at (4.25, 1.0) {$a$};
\node () at (4.25, 0.0) {$b$};
\draw [out=180, in=90] (3.5, 1.5) to (3.0, 1.25);
\draw [out=0, in=90] (3.5, 1.5) to (4.0, 1.25);
\draw [out=-90, in=90] (3.0, 1.25) to (3.0, 0.0);
\draw [out=-90, in=90] (4.0, 1.25) to (4.0, 0.75);
\draw [out=-90, in=90] (4.0, 0.25) to (4.0, 0.0);
\draw (3.75, 0.25) -- (4.25, 0.25) -- (4.25, 0.75) -- (3.75, 0.75) -- (3.75, 0.25);
\node () at (4.0, 0.5) {$\text{w}$};
\end{tikzpicture}

%% file: figures/tverbs.tex
\begin{tikzpicture}[scale=0.75]
\node () at (0.35, 1.0) {$n^r$};
\node () at (1.25, 1.0) {$s$};
\node () at (2.35, 1.0) {$n^l$};
\draw [out=-90, in=90] (0.0, 1.25) to (0.0, 0.0);
\draw [out=-90, in=90] (1.0, 1.25) to (1.0, 0.0);
\draw [out=-90, in=90] (2.0, 1.25) to (2.0, 0.0);
\draw (-0.25, 1.25) -- (2.25, 1.25) -- (1, 2) -- (-0.25, 1.25);
\node () at (1.0, 1.5) {$\text{v}$};
\node () at (3.0, 1.5) {$\mapsto$};
\node () at (4.35, 2.0) {$n^r$};
\node () at (5.25, 2.0) {$n$};
\node () at (6.25, 2.0) {$n$};
\node () at (7.35, 2.1) {$n^l$};
\node () at (5.75, 0.0) {$s$};
\draw [out=180, in=90] (4.5, 2.5) to (4.0, 2.25);
\draw [out=0, in=90] (4.5, 2.5) to (5.0, 2.25);
\draw [out=-90, in=90] (4.0, 2.25) to (4.0, 0.0);
\draw [out=-90, in=90] (5.0, 2.25) to (5.0, 1.25);
\draw [out=180, in=90] (6.5, 2.5) to (6.0, 2.25);
\draw [out=0, in=90] (6.5, 2.5) to (7.0, 2.25);
\draw [out=-90, in=90] (6.0, 2.25) to (6.0, 1.25);
\draw [out=-90, in=90] (7.0, 2.25) to (7.0, 0.0);
\draw [out=-90, in=90] (5.5, 0.75) to (5.5, 0.0);
\draw (4.75, 0.75) -- (6.25, 0.75) -- (6.25, 01.25) -- (4.75, 1.25) -- (4.75, 0.75);
\node () at (5.5, 1) {$\text{v}$};
\end{tikzpicture}

%% file: figures/relpron.tex
\begin{tikzpicture}
\node () at (0.25, 1.5) {$n^r$};
\node () at (1.25, 1.5) {$n$};
\node () at (2.25, 1.5) {$s^l$};
\node () at (3.25, 1.5) {$n$};
\draw [out=-90, in=90] (0.0, 1.75) to (0.0, 0.0);
\draw [out=-90, in=90] (1.0, 1.75) to (1.0, 0.0);
\draw [out=-90, in=90] (2.0, 1.75) to (2.0, 0.0);
\draw [out=-90, in=90] (3.0, 1.75) to (3.0, 0.0);
\draw (-0.25, 1.75) -- (1.5, 2.6) -- (3.25, 1.75) -- (-0.25, 1.75);
\node () at (1.5, 2.1) {who};
\node () at (4.0, 2.0) {$\mapsto$};
\node () at (5.25, 3.0) {$n^r$};
\node () at (7.75, 3.0) {$n$};
\node () at (6.25, 2.0) {$a$};
\node () at (9.25, 2.0) {$n$};
\node () at (7.25, 1.0) {$s$};
\node () at (8.25, 1.0) {$s^l$};
\node () at (6.75, 0.0) {$n$};
\draw (6.0, 3.5) .. controls (5.0, 3.5) .. (5.0, 3.25);
\draw (6.0, 3.5) .. controls (7.5, 3.5) .. (7.5, 3.25);
\draw (5.0, 3.25) .. controls (5.0, 0.0) .. (5.0, 0.0);
\draw (7.5, 3.25) .. controls (7.5, 2.75) .. (7.5, 2.75);
\draw (6.0, 2.25) .. controls (6.0, 0.75) .. (6.0, 0.75);
\draw (9.0, 2.25) .. controls (9.0, 0.0) .. (9.0, 0.0);
\draw (7.5, 1.5) .. controls (7.0, 1.5) .. (7.0, 1.25);
\draw (7.5, 1.5) .. controls (8.0, 1.5) .. (8.0, 1.25);
\draw (7.0, 1.25) .. controls (7.0, 0.75) .. (7.0, 0.75);
\draw (8.0, 1.25) .. controls (8.0, 0.0) .. (8.0, 0.0);
\draw (6.5, 0.25) .. controls (6.5, 0.0) .. (6.5, 0.0);
\draw (5.75, 2.25) -- (9.25, 2.25) -- (9.25, 2.75) -- (5.75, 2.75) -- (5.75, 2.25);
\node () at (7.5, 2.5) {$\text{who}_1$};
\draw (5.75, 0.25) -- (7.25, 0.25) -- (7.25, 0.75) -- (5.75, 0.75) -- (5.75, 0.25);
\node () at (6.5, 0.5) {$\text{who}_2$};
\end{tikzpicture}

%% file: figures/F_G_sentence.tex
\begin{tikzpicture}[rotate=90]
\begin{scope}[yscale=1.8]
\node () at (0.75, 8.0) {$d$};
\node () at (0.75, 7.0) {$n$};
\node () at (0.25, 6.0) {$a$};
\node () at (1.25, 6.0) {$n$};
\node () at (1.25, 5.0) {$s$};
\node () at (0.75, 4.0) {$n$};
\node () at (1.75, 3.0) {$d$};
\node () at (1.75, 2.0) {$n$};
\node () at (1.25, 1.0) {$s$};
\draw [out=-90, in=90] (0.5, 8.25) to (0.5, 7.75);
\draw [out=-90, in=90] (0.5, 7.25) to (0.5, 6.75);
\draw [out=-90, in=90] (0.0, 6.25) to (0.0, 4.75);
\draw [out=-90, in=90] (1.0, 6.25) to (1.0, 5.75);
\draw [out=-90, in=90] (1.0, 5.25) to (1.0, 4.75);
\draw [out=-90, in=90] (0.5, 4.25) to (0.5, 1.75);
\draw [out=-90, in=90] (1.5, 3.25) to (1.5, 2.75);
\draw [out=-90, in=90] (1.5, 2.25) to (1.5, 1.75);
\draw [out=-90, in=90] (1.0, 1.25) to (1.0, 0.75);
\draw (0.25, 8.25) -- (0.75, 8.25) -- (0.75, 8.75) -- (0.25, 8.75) -- (0.25, 8.25);
\node () at (0.5, 8.5) {the};
\draw (0.25, 7.25) -- (0.75, 7.25) -- (0.75, 7.75) -- (0.25, 7.75) -- (0.25, 7.25);
\node () at (0.5, 7.5) {person};
\draw (-0.25, 6.25) -- (1.25, 6.25) -- (1.25, 6.75) -- (-0.25, 6.75) -- (-0.25, 6.25);
\node () at (0.5, 6.5) {who$_1$};
\draw (0.75, 5.25) -- (1.25, 5.25) -- (1.25, 5.75) -- (0.75, 5.75) -- (0.75, 5.25);
\node () at (1.0, 5.5) {explains};
\draw (-0.25, 4.25) -- (1.25, 4.25) -- (1.25, 4.75) -- (-0.25, 4.75) -- (-0.25, 4.25);
\node () at (0.5, 4.5) {who$_2$};
\draw (1.25, 3.25) -- (1.75, 3.25) -- (1.75, 3.75) -- (1.25, 3.75) -- (1.25, 3.25);
\node () at (1.5, 3.5) {the};
\draw (1.25, 2.25) -- (1.75, 2.25) -- (1.75, 2.75) -- (1.25, 2.75) -- (1.25, 2.25);
\node () at (1.5, 2.5) {rules};
\draw (0.25, 1.25) -- (1.75, 1.25) -- (1.75, 1.75) -- (0.25, 1.75) -- (0.25, 1.25);
\node () at (1.0, 1.5) {knows};
\end{scope}
\end{tikzpicture}

%% file: 4-functorial-language.tex
We start this section by recasting the example of a language game modelled in
\cite{HedgesLewis18} into our formalism.
This is the first example given in the Philosophical Investigations
\cite{wittgenstein1953} and it features two players: a master builder and
his apprentice. The master gives instructions to the apprentice who
helps him building with building-stones.
They use a language consisting of the following dictionary
$D = \{ (\text{bring},\, s n^l) ,\; (\text{large},\, n n^l), \;
(\text{slabs},\, n),\: \dots\}\, ,$
where $n, s \in B$ are the \emph{noun} and \emph{sentence} type respectively.
The master gives an order to his apprentice by forming a grammatical sentence
$u \in O = \cal{L}(G, s)$ where $G = (B, V, D, s)$.
We can give a game-theoretic semantics to the master's orders as follows.
First, the monoidal signature associated with $G$ is
$\Gamma_G = \{ \text{bring}: n \to s, \: \text{large}: n \to n,
\: \text{slabs}: 1 \to n\}$,
and the functor $F_G$ maps ``Bring large slabs'' as follows:
\begin{equation}
  \begin{gathered}
    \input{figures/order.tex}
  \end{gathered}
  \quad \mapsto \quad \quad
  \begin{gathered}
    \input{figures/F_G_order.tex}
  \end{gathered}
\end{equation}
The interpretation as open games $J: \bf{MC}(\Gamma_G) \to \bf{Game}$ is given by $J(n) = N = \cal{L}(G, n)$ and $J(s) = \binom{O}{A}$ where $A$ is a
set of actions with a mapping $\tt{bring}: N \to A$, i.e. every noun phrase $x$ refers
to some object in the building site and $\tt{bring}(x)$ is the action that brings it.
We interpret the nouns and adjectives syntactically, i.e.
$J(\text{slabs}) = (\text{slabs} \to n) \in N$ and
\begin{equation*}\label{image-of-large}
  J(\text{large}) : \: N \longrightarrow N \: ::
  \begin{gathered}
    \input{figures/noun_var}
  \end{gathered}
  \quad \longmapsto \quad
  \begin{gathered}
    \input{figures/noun_phrase}
  \end{gathered}
  .
\end{equation*}
The game is encoded in the image of the imperative ``Bring'', which is
interpreted as the open game $\game{J(bring)}{N}{1}{O}{A}{}$ with trivial set
of strategy profiles $\Sigma = \star$ and play function:
$\Pf : \star \times N \to O: x \mapsto (id_s \otimes cup_n)
\circ (Bring \otimes x)\, ,$
analogous to the the image of ``large''. The coplay function is trivial
and the equilibrium
$\Ef : N \times (O \to A) \to \cal{P}(\star) = \bb{B}$ is given by:
\begin{equation}
  \Ef(x, k) = \begin{cases}
                1 & k(\Pf(x)) = \tt{bring}(x) \\
                0 & \text{otherwise}
              \end{cases}\, .
\end{equation}
Similar open games can be defined for other orders such as ``cut'' or ``stack''.
Applying the functor $\llbracket - \rrbracket = \cal{A}(J) \circ F_G$
to the diagram above, yields an open game
$\game{\llbracket Bring \, large \, slabs \rrbracket}{1}{1}{O}{A}{}$
with equilibrium function taking a \emph{continuation} of the apprentice
$k: O \to A$ to:
\begin{equation*}
  \Ef_{\llbracket Bring \, large \, slabs \rrbracket}(k) =
    \begin{cases}
        1 & k(\text{Bring large slabs}) = \tt{bring}(\text{large slabs})\\
        0 & \text{otherwise}
     \end{cases}
\end{equation*}
In words, the master is satisfied when his order results in an action that
brings large slabs.

Now that we have illustrated the concept of language game with a simple example,
we move to a different setup replacing the master-apprentice language
with teacher-student pragmatics. We start by modelling the role of a teacher
examining his pupil, where orders are replaced by questions.
Suppose we are dealing with a philosophy teacher and take
the following example:
\begin{gather*}
  D = \set{(\text{who},\, q s^l n \},\; (\text{invented},\, n^r s n^l),\;
      (\text{truth},\, nn^l),\; (\text{tables},\, n)}\\
  \input{figures/short_question.tex}
\end{gather*}
We model the teacher's knowledge of his course with a DisCo model
$K_T : \bf{G} \to \bf{Rel}$ such that $K_T(n) = K_T(q) = A$ where
$A = \{ \text{Fela}, \text{Wittgenstein}, \text{Peirce}, \dots \}$
is a set of names of historical figures.
The correct answers to a who-question $g: u \to q$ are given
by a subset $K_T(g) \subseteq A$, thus the utility of the student with strategy $k : Q \to A$ is the intersection $\langle k(g) \vert K_T(g) \rangle \in \bb{B}$, see example~\ref{example:queries}.

As in the previous example, we give a strategically-trivial semantics to
each of the words, except for the interrogative ``Who'' which we use to encode
the game. Note that the dictionary entry for ``Who'' factors via $F_G$ as
follows:
\begin{equation}
  \input{figures/who.tex}
\end{equation}
The functor $J: \bf{MC}(\Gamma_G) \to \bf{Game}$ is defined on objects by
$J(a) = J(n) = \binom{N}{1}$ and $J(q) = \binom{Q}{A}$ where
$Q = \cal{L}(G, q)$ is the set of grammatical who-questions and
$N = \cal{L}(G, n)$ is the set of grammatical noun phrases.
The first part of the factorization initializes the same noun variable
$x: 1 \to n \in \bf{G}$ in its two outputs:
$ \Pf_{J(\text{Who}_1)} = (x, x) \, : \, \star \to N \times N ,$
and has trivial coplay function, strategy set and equilibrium.
The second play function substitutes the question word ``Who'' for
$x$ in sentence $g$ to build the question:
$$ \Pf_{J(\text{Who}_2)}: N \times S \to Q: (x, g) \mapsto g[x := \text{Who}].$$
It has trivial coplay function and strategy set and an equilibrium
$\Ef_{J(\text{Who}_2)}: N \times S \times (Q \to A) \to \bb{B}$ given by
$\Ef_{J(\text{Who}_2)}(x, g, k) = \langle k(q) \vert K_T(q) \rangle$ for $q = \Pf_{J(\text{Who}_2)}(x, g)$.
As a result, the functor $\llbracket - \rrbracket = \cal{A}(J) \circ F_G$
maps the question ``Who invented truth tables?'' to an open game
with the following equilibrium function.
\begin{equation*}
  \Ef_{\llbracket \text{Who invented truth tables?} \rrbracket}(k) =
    \begin{cases}
        1 & k(\text{Who invented truth tables?}) \in
            \{ \text{Witt.}, \text{Peirce}\}\\
        0 & \text{otherwise}
     \end{cases}
\end{equation*}
In words, the teacher is satisfied when the student answers his question
correctly. Note that we could reverse the equilibrium to model a teacher
that is satisfied when the student gives the wrong answer.
In the next section, we will use this alternative choice to define an
adversarial question answering game.

%% file: figures/order.tex
\begin{tikzpicture}[scale=0.85]
\node () at (0.9166666666666666, -0.25) {};
\node () at (1.7, -0.25) {$n^l$};
\draw (0.0, 0) -- (2.0, 0) -- (1.0, 1) -- (0.0, 0);
\node () at (1.0, 0.25) {Bring};
\node () at (3.3, -0.35) {$n$};
\node () at (4.2, -0.25) {$n^l$};
\draw (2.5, 0) -- (4.5, 0) -- (3.5, 1) -- (2.5, 0);
\node () at (3.5, 0.25) {large};
\node () at (6.25, -0.25) {$n$};
\draw (5.0, 0) -- (7.0, 0) -- (6.0, 1) -- (5.0, 0);
\node () at (6.0, 0.25) {slabs};
\begin{scope}[yscale=0.8]
  \draw [out=-90, in=180] (1.3333333333333333, 0) to (2.25, -1);
  \draw [out=-90, in=0] (3.1666666666666665, 0) to (2.25, -1);
  \draw [out=-90, in=180] (3.833333333333333, 0) to (4.916666666666666, -1);
  \draw [out=-90, in=0] (6.0, 0) to (4.916666666666666, -1);
  \draw [out=-90, in=90] (0.6666666666666666, 0) to (0.6666666666666666, -2);
  \node () at (0.85, -0.45) {$s$};
\end{scope}
\end{tikzpicture}

%% file: figures/F_G_order.tex
\begin{tikzpicture}[scale=0.85]
\begin{scope}[xscale=1.3]
  \node () at (0.25, 2.0) {$n$};
  \node () at (0.25, 1.0) {$n$};
  \node () at (0.25, 0.0) {$s$};
  \draw [out=-90, in=90] (0.0, 2.25) to (0.0, 1.75);
  \draw [out=-90, in=90] (0.0, 1.25) to (0.0, 0.75);
  \draw [out=-90, in=90] (0.0, 0.25) to (0.0, 0);
  \draw (-0.5, 2.25) -- (0.5, 2.25) -- (0.5, 2.75) -- (-0.5, 2.75) -- (-0.5, 2.25);
  \node () at (0, 2.5) {slabs};
  \draw (-0.5, 1.25) -- (0.5, 1.25) -- (0.5, 1.75) -- (-0.5, 1.75) -- (-0.5, 1.25);
  \node () at (0.0, 1.5) {large};
  \draw (-0.5, 0.25) -- (0.5, 0.25) -- (0.5, 0.75) -- (-0.5, 0.75) -- (-0.5, 0.25);
  \node () at (0.0, 0.5) {Bring};
\end{scope}
\end{tikzpicture}

%% file: figures/noun_var.tex
\begin{tikzpicture}
\node () at (2.25, 0.0) {};
\draw [out=-90, in=90] (2.0, 0.25) to (2.0, -0.25);
\draw (1.5, 0.25) -- (2.5, 0.25) -- (2, 0.75) -- (1.5, 0.25);
\node () at (2, 0.4) {$x$};
\end{tikzpicture}

%% file: figures/noun_phrase.tex
\begin{tikzpicture}[scale=0.5]
\draw (-0.4, 0) -- (2.2, 0) -- (1, 1.3) -- (-0.4, 0);
\node () at (1.0, 0.25) {large};
\draw (2.7, 0) -- (4.3, 0) -- (3.5, 1) -- (2.7, 0);
\node () at (3.5, 0.25) {$x$};
\draw [out=-90, in=180] (1.3333333333333333, 0) to (2.4166666666666665, -1);
\draw [out=-90, in=0] (3.5, 0) to (2.4166666666666665, -1);
\draw [out=-90, in=90] (0.6666666666666666, 0) to (0.6666666666666666, -1.5);
\end{tikzpicture}

%% file: figures/short_question.tex
\begin{tikzpicture}[scale=0.75]
\draw (0.0, 0) -- (2.0, 0) -- (1.0, 1) -- (0.0, 0);
\node () at (1.0, 0.25) {Who};
\draw (2.5, 0) -- (4.5, 0) -- (3.5, 1) -- (2.5, 0);
\node () at (3.5, 0.25) {inv.};
\draw (5.0, 0) -- (7.0, 0) -- (6.0, 1) -- (5.0, 0);
\node () at (6.0, 0.25) {truth};
\draw (7.5, 0) -- (9.5, 0) -- (8.5, 1) -- (7.5, 0);
\node () at (8.5, 0.25) {tables};
\node () at (9.8, 0.25) {?};
\begin{scope}[yscale=0.8]
  \draw [out=-90, in=180] (1.5, 0) to (2.25, -1);
  \draw [out=-90, in=0] (3.0, 0) to (2.25, -1);
  \draw [out=-90, in=180] (4.0, 0) to (4.833333333333334, -1);
  \draw [out=-90, in=0] (5.666666666666667, 0) to (4.833333333333334, -1);
  \draw [out=-90, in=180] (6.333333333333333, 0) to (7.416666666666666, -1);
  \draw [out=-90, in=0] (8.5, 0) to (7.416666666666666, -1);
  \draw [out=-90, in=180] (1.0, 0) to (2.25, -2);
  \draw [out=-90, in=0] (3.5, 0) to (2.25, -2);
  \draw [out=-90, in=90] (0.5, 0) to (0.5, -2.3);
\end{scope}
\end{tikzpicture}

%% file: figures/who.tex
\begin{tikzpicture}[scale=0.85]
\node () at (1, 1.8) {$\text{Who}$};
\node () at (0.25, 1.3) {$q$};
\node () at (1.25, 1.3) {$s^l$};
\node () at (2.25, 1.3) {$n$};
\draw [out=-90, in=90] (0.0, 1.55) to (0.0, 0.0);
\draw [out=-90, in=90] (1.0, 1.55) to (1.0, 0.0);
\draw [out=-90, in=90] (2.0, 1.55) to (2.0, 0.0);
\draw (-0.25, 1.55) -- (2.25, 1.55) -- (1, 2.6) -- (-0.25, 1.55);
\node () at (1.0, 1.5) { };
\node () at (3.0, 1.5) {$\mapsto$};
\node () at (4.25, 2.0) {$a$};
\node () at (7.25, 2.0) {$n$};
\node () at (5.25, 1.0) {$s$};
\node () at (6.25, 1.0) {$s^l$};
\node () at (4.75, 0.0) {$q$};
\draw [out=-90, in=90] (4.0, 2.25) to (4.0, 0.75);
\draw [out=-90, in=90] (7.0, 2.25) to (7.0, 0.0);
\draw [out=180, in=90] (5.5, 1.5) to (5.0, 1.25);
\draw [out=0, in=90] (5.5, 1.5) to (6.0, 1.25);
\draw [out=-90, in=90] (5.0, 1.25) to (5.0, 0.75);
\draw [out=-90, in=90] (6.0, 1.25) to (6.0, 0.0);
\draw [out=-90, in=90] (4.5, 0.25) to (4.5, 0.0);
\draw (3.75, 2.25) -- (7.25, 2.25) -- (7.25, 2.75) -- (3.75, 2.75) -- (3.75, 2.25);
\node () at (5.5, 2.5) {$\text{Who}_1$};
\draw (3.75, 0.25) -- (5.25, 0.25) -- (5.25, 0.75) -- (3.75, 0.75) -- (3.75, 0.25);
\node () at (4.5, 0.5) {$\text{Who}_2$};
\end{tikzpicture}

%% file: 5-nash-equilibria.tex
In Section~\ref{sec:open-games}, we defined an open game for question-answering
using abstract sets $C$, $Q$, $A$ and $U$ for corpus, questions, answers and
utilities. Strategies and plays for the agents were given by arbitrary functions
between those sets. We now instantiate those functions with respect to a
pregroup grammar $G = (B, V, D, s)$ with a fixed question type $z \in B$.
We consider the case where utilities are booleans $U = \bb{B}$ and leave
the generalisation to any semiring for future work.

We take the corpus $C$ to be a list of question-answer pairs $(q, a)$ for
$q: u \to z$ and $a \in A$. For simplicity, we assume $q$ is a yes/no question
and $a$ is a boolean answer, i.e. $Q = \cal{L}(G, z)$ and $A = \bb{B}$.
The strategies of the student are DisCo models $\sigma: \bf{G} \to \bf{Rel}$
with $\sigma(z) = 1$, so that given a question $q: u \to z$,
$\sigma(q) \in \cal{P}(1) = \bb{B}$ is the student's answer. In practice,
the student may only have a subset of models available to him so  we set
$ \Sigma_\student \subseteq \{ \sigma: \bf{G} \to \bf{Rel}\, \colon\, \sigma(z) = 1\} $.
The strategies of the teacher are given by indices
$\Sigma_\teacher = \set{0, 1, \dots n}$, so that the play function
$\Pf_\teacher :  \Sigma_\teacher \times (Q \times A)^\ast \to Q \times A$
picks the question-answer pair indicated by the index.
The marker will compare the teacher's answer $a$ with the student's
answer $\sigma(q) \in \bb{B}$ using the metric
$\dist: A \times A \to \bb{B}:: (a_0, a_1) \mapsto (a_0 = a_1)$.
Plugging these open games as in Example~\ref{def:qa-game},
we can compute the set of equilibria of the game by composing the
equilibrium functions of its components.
\begin{equation*}
  \Ef_{\G}= \{ (j, \sigma) \in \Sigma_\teacher \times \Sigma_{\student}\,
  \colon\, j \in \amax_{i \in \Sigma_\teacher} a_{i} \neq \sigma(q_{i}) \land
  \sigma \in \amax_{\sigma \in \Sigma_\student} (a_{j} = \sigma(q_{j}))\}
\end{equation*}
Therefore, in a Nash equilibrium, the teacher will ask the question that the
student, even with his best guess, is going to answer in the worst way.
The student, on the other hand, is going to answer as correctly as possible.\\
We can analyse the possible outcomes of this game.
\begin{enumerate}
  \item There is a pair $(q_{i}, a_{i})$ in $C$ that the student cannot
    answer correctly, i.e. $ \forall \sigma \in \Sigma_\student\, \colon\, \sigma(q_{i}) \neq a_{i}$. Then $i$ is a winning strategy for the teacher
    and $(i, \sigma)$ is a Nash equilibrium, for any choice of strategy
    $\sigma$ for the student.
    If no such pair exists, then we fall into one of the following cases.
  \item The corpus is consistent --- i.e. $\exists \sigma: \bf{G} \to \bf{Rel}$
    such that $\forall i \cdot \sigma(q_i) = a_i$ --- and the student has access to the model $\sigma$ that answers all the possible questions correctly.
    Then, the strategy profile $(j, \sigma)$ is a Nash equilibrium and a
    winning strategy for the student for any choice $j$ of the teacher.
  \item For any choice $i$ of the teacher, the student has a model $\sigma_i$
    that answers $q_i$ correctly. And viceversa, for any strategy
    $\sigma$ of the student there is a choice $j$ of the teacher such that
    $\sigma(q_j) \neq a_j$. Then the set $\Ef_\G$ is empty, there is
    no Nash equilibrium.
\end{enumerate}

To illustrate the last case, consider a situation where the corpus $C = \set{(q_0, a_0), (q_1, a_1)}$ has only
two elements and the student has only
two models $\Sigma_\student = \set{\sigma_0, \sigma_1}$ such that
$\sigma_i(q_i) = a_i$ for $i \in \set{0, 1}$ but $\sigma_0(q_1) \neq a_1$ and
$\sigma_1(q_0) \neq a_0$. Then we're in a \emph{matching pennies} scenario,
both the teacher and the student have no incentive to choose any one of their
startegies and there is no Nash equilibrium. This problem can be ruled out if
we allowed the players in the game to have \emph{mixed strategies}, which can
be achieved with minor modifications of the open game formalism \cite{ghani2019}.

%% file: ms.bbl
\begin{thebibliography}{10}
\providecommand{\bibitemdeclare}[2]{}
\providecommand{\surnamestart}{}
\providecommand{\surnameend}{}
\providecommand{\urlprefix}{Available at }
\providecommand{\url}[1]{\texttt{#1}}
\providecommand{\href}[2]{\texttt{#2}}
\providecommand{\urlalt}[2]{\href{#1}{#2}}
\providecommand{\doi}[1]{doi:\urlalt{http://dx.doi.org/#1}{#1}}
\providecommand{\bibinfo}[2]{#2}

\bibitemdeclare{article}{benz2018}
\bibitem{benz2018}
\bibinfo{author}{Anton \surnamestart Benz\surnameend} \& \bibinfo{author}{Jon
  \surnamestart Stevens\surnameend} (\bibinfo{year}{2018}):
  \emph{\bibinfo{title}{Game-{{Theoretic Approaches}} to {{Pragmatics}}}}.
\newblock {\sl \bibinfo{journal}{Annual Review of Linguistics}}
  \bibinfo{volume}{4}(\bibinfo{number}{1}), pp. \bibinfo{pages}{173--191},
  \doi{{10.1146/annurev-linguistics-011817-045641}}.

\bibitemdeclare{inproceedings}{buszkowski2016}
\bibitem{buszkowski2016}
\bibinfo{author}{Wojciech \surnamestart Buszkowski\surnameend}
  (\bibinfo{year}{2016}): \emph{\bibinfo{title}{Syntactic {{Categories}} and
  {{Types}}: {{Ajdukiewicz}} and {{Modern Categorial Grammars}}}}.
\newblock \doi{10.1163/9789004311763\_004}.

\bibitemdeclare{inproceedings}{buszkowski2007}
\bibitem{buszkowski2007}
\bibinfo{author}{Wojciech \surnamestart Buszkowski\surnameend} \&
  \bibinfo{author}{Katarzyna \surnamestart Moroz\surnameend}
  (\bibinfo{year}{2007}): \emph{\bibinfo{title}{Pregroup {{Grammars}} and
  {{Context}}-Free {{Grammars}}}}.

\bibitemdeclare{inproceedings}{ClarkEtAl08}
\bibitem{ClarkEtAl08}
\bibinfo{author}{Stephen \surnamestart Clark\surnameend}, \bibinfo{author}{Bob
  \surnamestart Coecke\surnameend} \& \bibinfo{author}{Mehrnoosh \surnamestart
  Sadrzadeh\surnameend} (\bibinfo{year}{2008}): \emph{\bibinfo{title}{A
  {{Compositional Distributional Model}} of {{Meaning}}}}.
\newblock In: {\sl \bibinfo{booktitle}{Proceedings of the {{Second Symposium}}
  on {{Quantum Interaction}} ({{QI}}-2008)}}, pp. \bibinfo{pages}{133--140}.

\bibitemdeclare{incollection}{ClarkEtAl10}
\bibitem{ClarkEtAl10}
\bibinfo{author}{Stephen \surnamestart Clark\surnameend}, \bibinfo{author}{Bob
  \surnamestart Coecke\surnameend} \& \bibinfo{author}{Mehrnoosh \surnamestart
  Sadrzadeh\surnameend} (\bibinfo{year}{2010}):
  \emph{\bibinfo{title}{Mathematical Foundations for a Compositional
  Distributional Model of Meaning}}.
\newblock In \bibinfo{editor}{J.~\surnamestart {van Benthem}\surnameend},
  \bibinfo{editor}{M.~\surnamestart Moortgat\surnameend} \&
  \bibinfo{editor}{W.~\surnamestart Buszkowski\surnameend}, editors: {\sl
  \bibinfo{booktitle}{A {{Festschrift}} for {{Jim Lambek}}}}, {\sl
  \bibinfo{series}{Linguistic {{Analysis}}}}~\bibinfo{volume}{36}, pp.
  \bibinfo{pages}{345--384}.

\bibitemdeclare{article}{coecke2018b}
\bibitem{coecke2018b}
\bibinfo{author}{Bob \surnamestart Coecke\surnameend},
  \bibinfo{author}{Giovanni \surnamestart {de Felice}\surnameend},
  \bibinfo{author}{Dan \surnamestart Marsden\surnameend} \&
  \bibinfo{author}{Alexis \surnamestart Toumi\surnameend}
  (\bibinfo{year}{2018}): \emph{\bibinfo{title}{Towards {{Compositional
  Distributional Discourse Analysis}}}}.
\newblock {\sl \bibinfo{journal}{Electronic Proceedings in Theoretical Computer
  Science}} \bibinfo{volume}{283}, pp. \bibinfo{pages}{1--12},
  \doi{10.4204/EPTCS.283.1}.

\bibitemdeclare{article}{DeFeliceEtAl19a}
\bibitem{DeFeliceEtAl19a}
\bibinfo{author}{Giovanni \surnamestart {de Felice}\surnameend},
  \bibinfo{author}{Konstantinos \surnamestart Meichanetzidis\surnameend} \&
  \bibinfo{author}{Alexis \surnamestart Toumi\surnameend}
  (\bibinfo{year}{2019}): \emph{\bibinfo{title}{Functorial {{Question
  Answering}}}}.
\newblock {\sl
  \bibinfo{journal}{\href{https://arxiv.org/abs/1905.07408}{arXiv:1905.07408}
  [cs, math]}}.

\bibitemdeclare{article}{Delpeuch14a}
\bibitem{Delpeuch14a}
\bibinfo{author}{Antonin \surnamestart Delpeuch\surnameend}
  (\bibinfo{year}{2017}): \emph{\bibinfo{title}{Autonomization of {{Monoidal
  Categories}}}}.
\newblock \doi{10.31219/osf.io/efs3b}.

\bibitemdeclare{inproceedings}{fong2019}
\bibitem{fong2019}
\bibinfo{author}{Brendan \surnamestart Fong\surnameend}, \bibinfo{author}{David
  \surnamestart Spivak\surnameend} \& \bibinfo{author}{Remy \surnamestart
  Tuyeras\surnameend} (\bibinfo{year}{2019}): \emph{\bibinfo{title}{Backprop as
  Functor: A compositional perspective on supervised learning}}.
\newblock In: {\sl \bibinfo{booktitle}{2019 34th Annual {ACM}/{IEEE} Symposium
  on Logic in Computer Science ({LICS})}}, \bibinfo{publisher}{{IEEE}},
  \doi{10.1109/lics.2019.8785665}.

\bibitemdeclare{inproceedings}{fowler2008}
\bibitem{fowler2008}
\bibinfo{author}{Timothy A.~D. \surnamestart Fowler\surnameend}
  (\bibinfo{year}{2008}): \emph{\bibinfo{title}{Efficiently {{Parsing}} with
  the {{Product}}-{{Free Lambek Calculus}}}}.
\newblock In: {\sl \bibinfo{booktitle}{Proceedings of the 22nd {{International
  Conference}} on {{Computational Linguistics}} ({{Coling}} 2008)}},
  \bibinfo{publisher}{{Coling 2008 Organizing Committee}},
  \bibinfo{address}{{Manchester, UK}}, pp. \bibinfo{pages}{217--224},
  \doi{10.3115/1599081.1599109}.

\bibitemdeclare{inproceedings}{GhaniHedges18}
\bibitem{GhaniHedges18}
\bibinfo{author}{Neil \surnamestart Ghani\surnameend}, \bibinfo{author}{Jules
  \surnamestart Hedges\surnameend}, \bibinfo{author}{Viktor \surnamestart
  Winschel\surnameend} \& \bibinfo{author}{Philipp \surnamestart
  Zahn\surnameend} (\bibinfo{year}{2018}): \emph{\bibinfo{title}{Compositional
  game theory}}.
\newblock In: {\sl \bibinfo{booktitle}{Proceedings of the 33rd Annual ACM/IEEE
  Symposium on Logic in Computer Science}}, pp. \bibinfo{pages}{472--481},
  \doi{10.3982/ECTA6297}.

\bibitemdeclare{inproceedings}{ghani2019}
\bibitem{ghani2019}
\bibinfo{author}{Neil \surnamestart Ghani\surnameend}, \bibinfo{author}{Clemens
  \surnamestart Kupke\surnameend}, \bibinfo{author}{Alasdair \surnamestart
  Lambert\surnameend} \& \bibinfo{author}{Fredrik~Nordvall \surnamestart
  Forsberg\surnameend} (\bibinfo{year}{2019}):
  \emph{\bibinfo{title}{Compositional Game Theory with Mixed Strategies:
  Probabilistic Open Games Using a Distributive Law}}.
\newblock In: {\sl \bibinfo{booktitle}{Applied Category Theory Conference
  2019}}.

\bibitemdeclare{incollection}{goodfellow2014a}
\bibitem{goodfellow2014a}
\bibinfo{author}{Ian \surnamestart Goodfellow\surnameend},
  \bibinfo{author}{Jean \surnamestart {Pouget-Abadie}\surnameend},
  \bibinfo{author}{Mehdi \surnamestart Mirza\surnameend}, \bibinfo{author}{Bing
  \surnamestart Xu\surnameend}, \bibinfo{author}{David \surnamestart
  {Warde-Farley}\surnameend}, \bibinfo{author}{Sherjil \surnamestart
  Ozair\surnameend}, \bibinfo{author}{Aaron \surnamestart Courville\surnameend}
  \& \bibinfo{author}{Yoshua \surnamestart Bengio\surnameend}
  (\bibinfo{year}{2014}): \emph{\bibinfo{title}{Generative {{Adversarial
  Nets}}}}.
\newblock In \bibinfo{editor}{Z.~\surnamestart Ghahramani\surnameend},
  \bibinfo{editor}{M.~\surnamestart Welling\surnameend},
  \bibinfo{editor}{C.~\surnamestart Cortes\surnameend}, \bibinfo{editor}{N.~D.
  \surnamestart Lawrence\surnameend} \& \bibinfo{editor}{K.~Q. \surnamestart
  Weinberger\surnameend}, editors: {\sl \bibinfo{booktitle}{Advances in
  {{Neural Information Processing Systems}} 27}}, \bibinfo{publisher}{{Curran
  Associates, Inc.}}, pp. \bibinfo{pages}{2672--2680}.

\bibitemdeclare{inproceedings}{Grefenstette11}
\bibitem{Grefenstette11}
\bibinfo{author}{Edward \surnamestart Grefenstette\surnameend} \&
  \bibinfo{author}{Mehrnoosh \surnamestart Sadrzadeh\surnameend}
  (\bibinfo{year}{2011}): \emph{\bibinfo{title}{Experimental {{Support}} for a
  {{Categorical Compositional Distributional Model}} of {{Meaning}}}}.
\newblock In: {\sl \bibinfo{booktitle}{The 2014 {{Conference}} on {{Empirical
  Methods}} on {{Natural Language Processing}}.}}, pp.
  \bibinfo{pages}{1394--1404}.

\bibitemdeclare{article}{Hedges17}
\bibitem{Hedges17}
\bibinfo{author}{Jules \surnamestart Hedges\surnameend} (\bibinfo{year}{2017}):
  \emph{\bibinfo{title}{Coherence for Lenses and Open Games}}.
\newblock {\sl
  \bibinfo{journal}{\href{https://arxiv.org/abs/1704.02230}{arXiv:1704.02230}
  [cs, math]}}.

\bibitemdeclare{article}{hedges2019}
\bibitem{hedges2019}
\bibinfo{author}{Jules \surnamestart Hedges\surnameend} (\bibinfo{year}{2019}):
  \emph{\bibinfo{title}{From Open Learners to Open Games}}.
\newblock {\sl
  \bibinfo{journal}{\href{https://arxiv.org/abs/1902.08666}{arXiv:1902.08666}
  [cs, math]}}.

\bibitemdeclare{article}{HedgesLewis18}
\bibitem{HedgesLewis18}
\bibinfo{author}{Jules \surnamestart Hedges\surnameend} \&
  \bibinfo{author}{Martha \surnamestart Lewis\surnameend}
  (\bibinfo{year}{2018}): \emph{\bibinfo{title}{Towards Functorial
  Language-Games}}.
\newblock {\sl \bibinfo{journal}{Electronic Proceedings in Theoretical Computer
  Science}} \bibinfo{volume}{283}, pp. \bibinfo{pages}{89--102},
  \doi{10.4204/eptcs.283.7}.

\bibitemdeclare{inproceedings}{kissinger2017}
\bibitem{kissinger2017}
\bibinfo{author}{Aleks \surnamestart Kissinger\surnameend} \&
  \bibinfo{author}{Sander \surnamestart Uijlen\surnameend}
  (\bibinfo{year}{2017}): \emph{\bibinfo{title}{A categorical semantics for
  causal structure}}.
\newblock In: {\sl \bibinfo{booktitle}{32nd Annual {ACM/IEEE} Symposium on
  Logic in Computer Science, {LICS} 2017, Reykjavik, Iceland, June 20-23,
  2017}}, \bibinfo{publisher}{{IEEE} Computer Society}, pp.
  \bibinfo{pages}{1--12}, \doi{10.1109/LICS.2017.8005095}.

\bibitemdeclare{inproceedings}{Lambek99}
\bibitem{Lambek99}
\bibinfo{author}{Joachim \surnamestart Lambek\surnameend}
  (\bibinfo{year}{1999}): \emph{\bibinfo{title}{Type {{Grammar Revisited}}}}.
\newblock In \bibinfo{editor}{Alain \surnamestart Lecomte\surnameend},
  \bibinfo{editor}{Fran{\c c}ois \surnamestart Lamarche\surnameend} \&
  \bibinfo{editor}{Guy \surnamestart Perrier\surnameend}, editors: {\sl
  \bibinfo{booktitle}{Logical {{Aspects}} of {{Computational Linguistics}}}},
  \bibinfo{publisher}{{Springer Berlin Heidelberg}}, \bibinfo{address}{{Berlin,
  Heidelberg}}, pp. \bibinfo{pages}{1--27}, \doi{10.1016/0168-0072(94)00063-9}.

\bibitemdeclare{book}{Lambek08}
\bibitem{Lambek08}
\bibinfo{author}{Joachim \surnamestart Lambek\surnameend}
  (\bibinfo{year}{2008}): \emph{\bibinfo{title}{From {{Word}} to {{Sentence}}:
  {{A Computational Algebraic Approach}} to {{Grammar}}}}.
\newblock \bibinfo{series}{Open Access Publications},
  \bibinfo{publisher}{{Polimetrica}}.

\bibitemdeclare{book}{lewis1969}
\bibitem{lewis1969}
\bibinfo{author}{David~K. \surnamestart Lewis\surnameend}
  (\bibinfo{year}{1969}): \emph{\bibinfo{title}{Convention: {{A Philosophical
  Study}}}}.
\newblock \bibinfo{publisher}{{Wiley-Blackwell}}.

\bibitemdeclare{article}{Montague70a}
\bibitem{Montague70a}
\bibinfo{author}{Richard \surnamestart Montague\surnameend}
  (\bibinfo{year}{1970}): \emph{\bibinfo{title}{Universal Grammar}}.
\newblock {\sl \bibinfo{journal}{Theoria}}
  \bibinfo{volume}{36}(\bibinfo{number}{3}), pp. \bibinfo{pages}{373--398},
  \doi{10.1111/j.1755-2567.1970.tb00434.x}.

\bibitemdeclare{article}{preller2007}
\bibitem{preller2007}
\bibinfo{author}{Anne \surnamestart Preller\surnameend} (\bibinfo{year}{2007}):
  \emph{\bibinfo{title}{Linear {{Processing}} with {{Pregroups}}}}.
\newblock {\sl \bibinfo{journal}{Studia Logica}}
  \bibinfo{volume}{87}(\bibinfo{number}{2-3}), pp. \bibinfo{pages}{171--197},
  \doi{10.1007/s11225-007-9087-0}.

\bibitemdeclare{article}{PrellerLambek07}
\bibitem{PrellerLambek07}
\bibinfo{author}{Anne \surnamestart Preller\surnameend} \&
  \bibinfo{author}{Joachim \surnamestart Lambek\surnameend}
  (\bibinfo{year}{2007}): \emph{\bibinfo{title}{Free Compact 2-Categories}}.
\newblock {\sl \bibinfo{journal}{Mathematical Structures in Computer Science}}
  \bibinfo{volume}{17}(\bibinfo{number}{2}), pp. \bibinfo{pages}{309--340},
  \doi{10.1017/S0960129506005901}.

\bibitemdeclare{article}{roman2020}
\bibitem{roman2020}
\bibinfo{author}{Mario \surnamestart Rom{\'a}n\surnameend}
  (\bibinfo{year}{2020}): \emph{\bibinfo{title}{Comb {{Diagrams}} for
  {{Discrete}}-{{Time Feedback}}}}.
\newblock {\sl
  \bibinfo{journal}{\href{https://arxiv.org/abs/2003.06214}{arXiv:2003.06214}
  [cs]}}.

\bibitemdeclare{article}{sadrzadeh2013}
\bibitem{sadrzadeh2013}
\bibinfo{author}{Mehrnoosh \surnamestart Sadrzadeh\surnameend},
  \bibinfo{author}{Stephen \surnamestart Clark\surnameend} \&
  \bibinfo{author}{Bob \surnamestart Coecke\surnameend} (\bibinfo{year}{2013}):
  \emph{\bibinfo{title}{The {{Frobenius}} Anatomy of Word Meanings {{I}}:
  Subject and Object Relative Pronouns}}.
\newblock {\sl \bibinfo{journal}{Journal of Logic and Computation}}
  \bibinfo{volume}{23}(\bibinfo{number}{6}), pp. \bibinfo{pages}{1293--1317},
  \doi{10.1093/logcom/ext044}.

\bibitemdeclare{incollection}{Selinger09}
\bibitem{Selinger09}
\bibinfo{author}{P.~\surnamestart Selinger\surnameend} (\bibinfo{year}{2010}):
  \emph{\bibinfo{title}{A {{Survey}} of {{Graphical Languages}} for {{Monoidal
  Categories}}}}.
\newblock In: {\sl \bibinfo{booktitle}{New {{Structures}} for {{Physics}}}},
  \bibinfo{publisher}{{Springer Berlin Heidelberg}}, pp.
  \bibinfo{pages}{289--355}, \doi{10.1007/978-3-642-12821-9\_4}.

\bibitemdeclare{book}{steedman2000}
\bibitem{steedman2000}
\bibinfo{author}{Mark \surnamestart Steedman\surnameend}
  (\bibinfo{year}{2000}): \emph{\bibinfo{title}{The Syntactic Process}}.
\newblock \bibinfo{publisher}{{MIT Press}}, \bibinfo{address}{{Cambridge, MA,
  USA}}.

\bibitemdeclare{inproceedings}{subramanian2017}
\bibitem{subramanian2017}
\bibinfo{author}{Sandeep \surnamestart Subramanian\surnameend},
  \bibinfo{author}{Sai \surnamestart Rajeswar\surnameend},
  \bibinfo{author}{Francis \surnamestart Dutil\surnameend},
  \bibinfo{author}{Chris \surnamestart Pal\surnameend} \&
  \bibinfo{author}{Aaron \surnamestart Courville\surnameend}
  (\bibinfo{year}{2017}): \emph{\bibinfo{title}{Adversarial {{Generation}} of
  {{Natural Language}}}}.
\newblock In: {\sl \bibinfo{booktitle}{Proceedings of the 2nd {{Workshop}} on
  {{Representation Learning}} for {{NLP}}}}, \bibinfo{publisher}{{Association
  for Computational Linguistics}}, \bibinfo{address}{{Vancouver, Canada}}, pp.
  \bibinfo{pages}{241--251}, \doi{10.18653/v1/W17-2629}.

\bibitemdeclare{inproceedings}{tripodi2019}
\bibitem{tripodi2019}
\bibinfo{author}{Rocco \surnamestart Tripodi\surnameend} \&
  \bibinfo{author}{Roberto \surnamestart Navigli\surnameend}
  (\bibinfo{year}{2019}): \emph{\bibinfo{title}{Game {{Theory Meets
  Embeddings}}: A {{Unified Framework}} for {{Word Sense Disambiguation}}}}.
\newblock In: {\sl \bibinfo{booktitle}{Proceedings of the 2019 {{Conference}}
  on {{Empirical Methods}} in {{Natural Language Processing}} and the 9th
  {{International Joint Conference}} on {{Natural Language Processing}}
  ({{EMNLP}}-{{IJCNLP}})}}, \bibinfo{publisher}{{Association for Computational
  Linguistics}}, \bibinfo{address}{{Hong Kong, China}}, pp.
  \bibinfo{pages}{88--99}, \doi{10.18653/v1/D19-1009}.

\bibitemdeclare{book}{wittgenstein1953}
\bibitem{wittgenstein1953}
\bibinfo{author}{Ludwig \surnamestart Wittgenstein\surnameend}
  (\bibinfo{year}{1953}): \emph{\bibinfo{title}{Philosophical Investigations}}.
\newblock \bibinfo{publisher}{Basil Blackwell}, \bibinfo{address}{Oxford}.

\end{thebibliography}
